\documentclass[runningheads]{llncs}
\usepackage{graphicx}
\usepackage{amsmath}
\usepackage{multirow}
\usepackage{xcolor}
\usepackage{subfig}
\usepackage{booktabs} 
\usepackage{array}
\usepackage{caption}
\usepackage{cite}
\usepackage{hyperref}
\usepackage{footmisc}
\begin{document}

\title{Mind the Gap:\\Analyzing Lacunae with \\ Transformer-Based Transcription}

%\title{Contribution Title\thanks{Supported by organization x.}}

\author{Jaydeep Borkar \and
David A. Smith\orcidID{0000-0002-6636-6940}}

\authorrunning{Borkar and Smith}
\institute{Khoury College of Computer Sciences\\
Northeastern University\\
Boston, MA 02115, U.S.A\\
\email{borkar.j@northeastern.edu}, \email{dasmith@ccs.neu.edu}}

\maketitle              % typeset the header of the contribution
\renewcommand{\thefootnote}{}
\footnotetext{Accepted to the ICDAR 2024 \textit{Workshop on Computational Paleography,} Athens, Greece, 2024.}

\begin{abstract}
Historical documents frequently suffer from damage and inconsistencies, including missing or illegible text resulting from issues such as holes, ink problems, and storage damage. These missing portions or gaps are referred to as lacunae. In this study, we employ transformer-based optical character recognition (OCR) models trained on synthetic data containing lacunae in a supervised manner. We demonstrate their effectiveness in detecting and restoring lacunae, achieving a success rate of 65\%, compared to a base model lacking knowledge of lacunae, which achieves only 5\% restoration. Additionally, we investigate the mechanistic properties of the model, such as the log probability of transcription, which can identify lacunae and other errors (e.g., mistranscriptions due to complex writing or ink issues) in line images without directly inspecting the image. This capability could be valuable for scholars seeking to distinguish images containing lacunae or errors from clean ones. Although we explore the potential of attention mechanisms in flagging lacunae and transcription errors, our findings suggest it is not a significant factor. Our work highlights a promising direction in utilizing transformer-based OCR models for restoring or analyzing damaged historical documents.

\end{abstract}

\section{Introduction}
\label{sec:intro}

Many historical documents have reached us in incomplete form. Besides the larger losses of missing leaves or inscribed surfaces, existing pages and other carriers of writing often exhibit lacunae, gaps at the edge or in the middle of lines, or missing lines in the middle of a text. Moreover, when working with digital surrogates for historical documents, we may encounter gaps introduced by the imaging process, such as overly dark microfilm or poorly cropped photographs. Paleography, papyrology, and epigraphy have developed standards, such as the Leiden conventions, for distinguishing readable text, partial letters, abbreviated forms, and restored lacunae \cite{van_groningen_signis_1932}.

Building on progress in handwritten text recognition (HTR) \cite{nockels_implications_2024} and language modeling \cite{devlin-etal-2019-bert}, researchers have employed language technologies to good effect in the traditional philological task of hypothesizing readings for these lacunae \emph{when their locations are known} \cite{assael_restoring_2022,cowen-breen_logion_2023}.  If asked to transcribe a new document, however, these restoration models do not know where any gaps in them might be.  Furthermore, as language technologies become more powerful, transcription models may suggest readings that are not supported by input document images (or, similarly, sound recordings \cite{koenecke_careless_2024}).

This paper considers two related inference problems that arise when HTR models are confronted with images of a document that may contain lacunae.  First, is the HTR model capable of accurately filling the gap left by the absence of visual evidence of writing?  Second, does the model provide information about the presence of lacunae, so that users are aware of which parts of a transcript are inferred from context alone?  We address the first question by measuring HTR transcription accuracy on lines with and without lacunae, evaluating models trained with various proportions of lacunae.  We address the second question by training models to predict the presence of lacunae and other errors in text lines at inference time.

As in many other areas of natural language processing, transformer architectures are being utilized in OCR.  We perform experimental evaluations with the TrOCR model, which combines a pre-trained vision transformer with a pre-trained language model decoder \cite{trocr}.  Besides state-of-the-art performance, transformer architectures employ an attention mechanism, which provides us with a way to probe the relationship between visual evidence and transcription output.  To control the prevalence of lacunae experimentally, we identify the presence of individual characters in the training and test data using OCR and randomly assign some proportion of them to be missing.

This study reaches the following main findings: 
\begin{itemize}
    \item Transformer models like TrOCR that are pre-trained on clean text struggle to transfer their knowledge effectively for inferring (the likely content of) lacunae in line images.
    \item Adding examples with lacunae to TrOCR's supervised training significantly enhances its rate of lacuna restoration from 5.6\% to 65.85\%.
    \item Using TrOCR's log probability of an HTR transcript as a predictor in a logistic regression model correctly flags lines containing lacunae 53\% of the time and images with other errors 84\% of the time. 
    \item Using a sample of the model's attention weights to predict lacunae in input images and mistranscriptions in the output achieves accuracy not significantly higher than what was already reached using the log probability of the transcription. Nonetheless, we believe further experiments probing HTR models' representations of their input would be useful.
\end{itemize}

After reviewing related work (\S\ref{sec:related}), we describe the construction of the training and test data for our study and the inference procedures used in our experiments (\S\ref{sec:approach}).  We then describe the experimental setup (\S\ref{sec:experiments}) and discuss the results (\S\ref{sec:results}).  We conclude with a discussion of future directions for research on lacunose manuscripts (\S\ref{sec:conclusion}).

\section{Related Work}
\label{sec:related}

Efforts have been made to restore lacunae through a pretrain-finetune framework \cite{vogler-etal-2022-lacuna}. In this approach, authors utilized self-supervised contrastive loss training, followed by supervised fine-tuning. This method showed significant improvement in the restoration of lacunae within printed and handwritten documents, both in English and Arabic languages. There have been numerous works in NLP in the domain of masked pretraining where the objective is to learn the masked representations from unlabeled data \cite{devlin-etal-2019-bert}. In the speech domain, Baevski et al. proposed wav2vec 2.0 framework that masks the speech input and solves a contrastive task over quantized speech representations by using self-supervised contrastive loss during pre-training and Connectionist Temporal Classification (CTC) loss during fine-tuning \cite{NEURIPS2020_92d1e1eb}. Damage and inconsistencies in historical documents, such as inking issues, often pose challenges when using modern OCR models for transcription \cite{ArlitschHerbert+2004+59+67}. Several OCR post-correction models have been proposed to correct the transcription of OCR models when dealing with historical documents\cite{ocr1, dong-smith-2018-multi}, including some for endangered languages\cite{rijhwani-etal-2020-ocr, rijhwani-etal-2021-lexically}.

Transformer-based models have been widely used in language modeling tasks \cite{radford2019language, devlin-etal-2019-bert, t5} and in optical character recognition to transcribe the text present in the image\cite{trocr}. Transformer-based OCR models, such as TrOCR, use an encoder-decoder framework. In this framework, a pre-trained image transformer serves as the encoder, while a pre-trained text transformer acts as the decoder. TrOCR has demonstrated superior performance over current state-of-the-art models across various domains, including printed, handwritten, and scene text recognition tasks.

\section{Approach}
\label{sec:approach}

\subsection{Dataset Creation}
\label{sec:data}

In this study, we create synthetic lacunae by utilizing line images from the IAM handwriting database\footnote{https://fki.tic.heia-fr.ch/databases/iam-handwriting-database}, replicating real-world scenarios. We begin by extracting all bounding boxes along with their coordinates from the line image using Pytesseract\footnote{https://github.com/tesseract-ocr/tesseract}. Next, we randomly select a bounding box from the list and apply binary thresholding to that region with a threshold value of 10, rendering it completely white to create the lacuna.  Gaps generated in this way resemble the lacunae introduced into binarized microfilm images.

Subsequently, we binarize the remaining portions of the line image to improve quality for optical character recognition. We explore two approaches for binarizing line images: static thresholding, which applies the same threshold to all images, and adaptive thresholding, which dynamically adjusts the threshold value for each image based on its pixel distribution and layout. Our findings reveal that static thresholding adversely affects the visual quality of non-lacuna characters. Therefore, we opt for adaptive thresholding, which preserves the visual integrity of non-lacuna characters. Consequently, we utilize adaptively thresholded binarized images for lacuna creation in our experiments. 

\begin{figure}[htbp]
    \centering
    \subfloat[Ground truth: nominating any more Labour life Peers.]{%
        \includegraphics[width=0.45\textwidth]{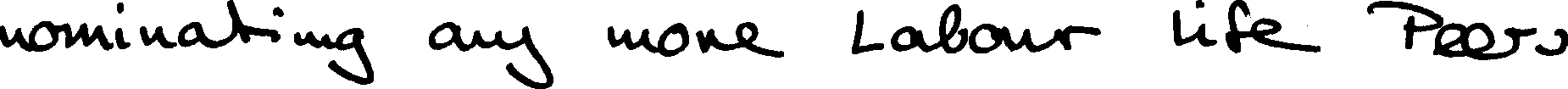}%
        \label{fig:sub1}%
    }\hfill
    \subfloat[nominating any \textbf{more} Labour life Peers.]{%
        \includegraphics[width=0.45\textwidth]{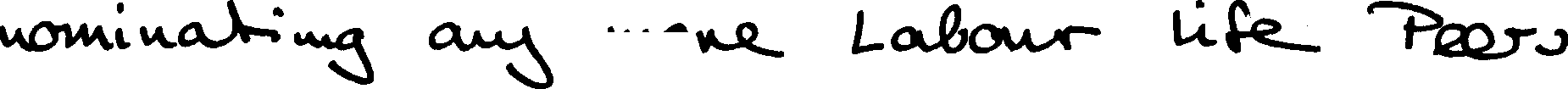}%
        \label{fig:sub2}%
    }
    
    \subfloat[Ground truth: Lancaster House despite the crisis which had.]{%
        \includegraphics[width=0.45\textwidth]{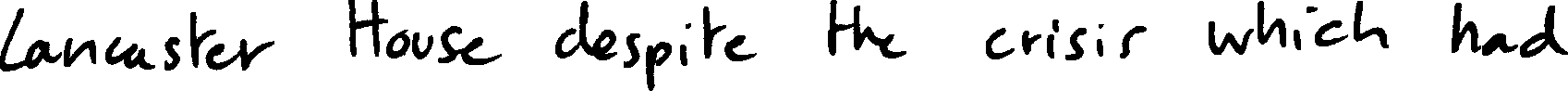}%
        \label{fig:sub3}%
    }\hfill
    \subfloat[Lancaster House despite the \textbf{crisis} which had]{%
        \includegraphics[width=0.45\textwidth]{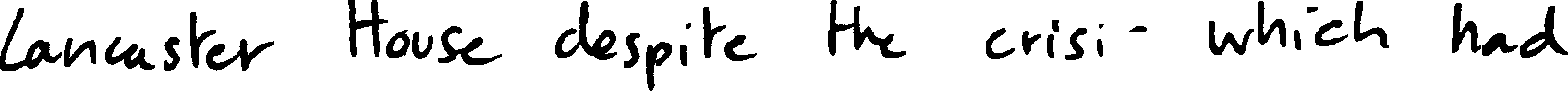}%
        \label{fig:sub4}%
    }
    
    \subfloat[Ground truth: round a doll's house.]{%
        \includegraphics[width=0.45\textwidth]{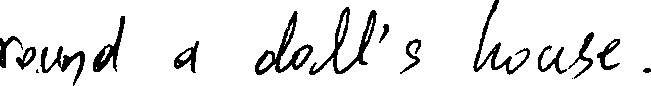}%
        \label{fig:sub5}%
    }\hfill
    \subfloat[round a doll's \textbf{house}.]{%
        \includegraphics[width=0.45\textwidth]{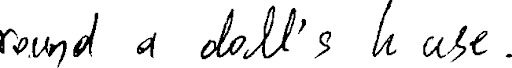}%
        \label{fig:sub6}%
    }
    
    \caption{Line images (b, d, f) containing lacuna highlighted in bold.}
    \label{fig:lacuna}
\end{figure}

We also process the ground-truth labels to address certain irregularities. Initially, characters of words that were originally separated by white spaces are recombined (e.g., ``B B C'' becomes ``BBC''). Additionally, contractions that were inconsistently tokenized from the original data, separating them from the words, are reattached to the respective words (e.g., ``He 'll'' becomes ``He'll'', ``They 've'' becomes ``They've''). Figure~\ref{fig:lacuna} shows some samples of our line images containing lacunae.

\subsection{Inference and Evaluation}
\label{sec:inference}

To evaluate the performance of models on our datasets, we use the Character Error Rate (CER), a metric that is widely used to evaluate the effectiveness of optical character recognition. It measures the rate of character-level errors in the OCR output compared to the ground truth text. CER is given by 

\[
\text{CER} = \frac{N_{\text{insertions}} + N_{\text{deletions}} + N_{\text{substitutions}}}{N}
\]

\noindent where \textit{N} is the total number of characters in the ground truth,  $N_{\text{insertions}}$ is the number of characters incorrectly inserted by the model, $N_{\text{deletions}}$ is the number of characters incorrectly deleted by the model, and $N_{\text{substitutions}}$ is the number of characters incorrectly substituted by the model. Let $X$ represent the line image, $S$ denote its text sequence, $n$ signify the total number of words in $S$, $L$ denote the lacuna token, and $\hat{L}$ symbolize the model's predicted lacuna token. With $S = (w_1, w_2, ..., w_n)$ as the text sequence comprising $n$ words, we assess whether $\hat{L} = L$ when $L$ appears at position $n_i$ in $S$.

\subsection{Baseline: Effects of Lacunae and Complex Writing on Log Probabilities}
\label{section:baseline}

We aim to investigate the impact of lacunae and other factors, such as characters or words written in an unusual manner, on the log probability of transcription tokens. Log probability is a measure used to quantify the likelihood of a word or sequence of words occurring given some context. In the context of OCR, it represents the probability of predicting a new token given previously transcribed tokens. Let $w_1, w_2, ..., w_n$ be a sequence of $n$ tokens in the transcription, $w_{1:(i-1)}$ be the sequence of tokens from the first token up to $w_i$, $P(w_i | w_{1:(i-1)})$ be the conditional probability of observing token $w_i$ given the context $w_{1:(i-1)}$. The log probability of a token $w_i$ given its context $w_{1:(i-1)}$ is:

\[\text{Log Probability} = \log P(w_i | w_{1:(i-1)})\]

Our hypothesis suggests that the log probability would decrease for tokens affected by lacunae, as the model would exhibit uncertainty in its interpretation. Similarly, this hypothesis extends to tokens written in a complex manner, where the model may also struggle to accurately predict them thus resulting in lower log probabilities. To test our hypothesis, we train a logistic regression model using the minimum of the log probability (\textit{\( \text{min}(\text{log probability}) \)}) as a feature. The \textit{\( \text{min}(\text{log probability}) \)} represents the lowest log probability observed among all transcription tokens. Let \( P_{\text{min}} \) be the minimum of all log probabilities in the transcription tokens, then the logistic regression model can be represented as 
\[
\text{Logistic Regression: } y = \sigma(\beta_0 + \beta_1 \cdot P_{\text{min}})
\]

\noindent where  $y$ represents the predicted outcome (e.g., lacuna or error detection), $\beta_0$ is the intercept, \( \beta_1 \) is the coefficient associated with the feature \( P_{\text{min}} \), and \( \sigma \) is the logistic function. We report our findings in \S\ref{section:attention}.

\subsection{Hypothesis: Can Attention Detect Lacunae and Complex Writing?} 

In models such as TrOCR, cross-attention refers to the mechanism by which the models attend to different parts of the input line image while generating the output transcription. Cross-attention helps the model focus on the relevant regions of the input image at each step of the transcription process, enabling it to capture any contextual dependencies for accurate optical character recognition. Given an input image $I$ consisting of a sequence of visual features $\{v_1, v_2, ..., v_n\}$ representing different regions in the image, and a target transcription $T$ consisting of a sequence of tokens $\{t_1, t_2, ..., t_m\}$ representing the characters in the text, the cross-attention mechanism computes a set of attention scores $\{\alpha_1, \alpha_2, ..., \alpha_n\}$, where each attention score $\alpha_i$ indicates the relevance of the $i^{th}$ visual feature to the generation of the current token $t_j$ in the transcription.

We study a possible hypothesis of identifying any attention patterns in the cross-attention matrix to visually identify lacunae and complex writing. Our hypothesis suggests that when the model processes lacuna characters or complex handwriting lacking sufficient visual evidence, the model may allocate greater attention scores to other regions within the line image that provide clearer visual cues. Further, it can utilize this contextual information to infer the identity of the lacuna or other characters written in a complex way. To test this hypothesis, we train a logistic regression model using two features: \textit{\( \text{min}(\text{log probability})\)} (as in section \ref{section:baseline}) and \textit{\( \text{attention entropy}\)}. \textit{\( \text{attention entropy}\)} is a measure used to quantify the uncertainty or randomness in the distribution of attention scores. Entropy $H$ for an attention distribution $p$ can be given as 

\vspace{0.25em} 
\begin{center}
    Entropy \(H(p) = -\sum_{i=1}^{n} p_i \log(p_i)\) \\ 
\end{center}
\vspace{0.25em} 

\begin{figure}
    \centering
    \subfloat[Log Probability]{\includegraphics[width=0.45\textwidth]{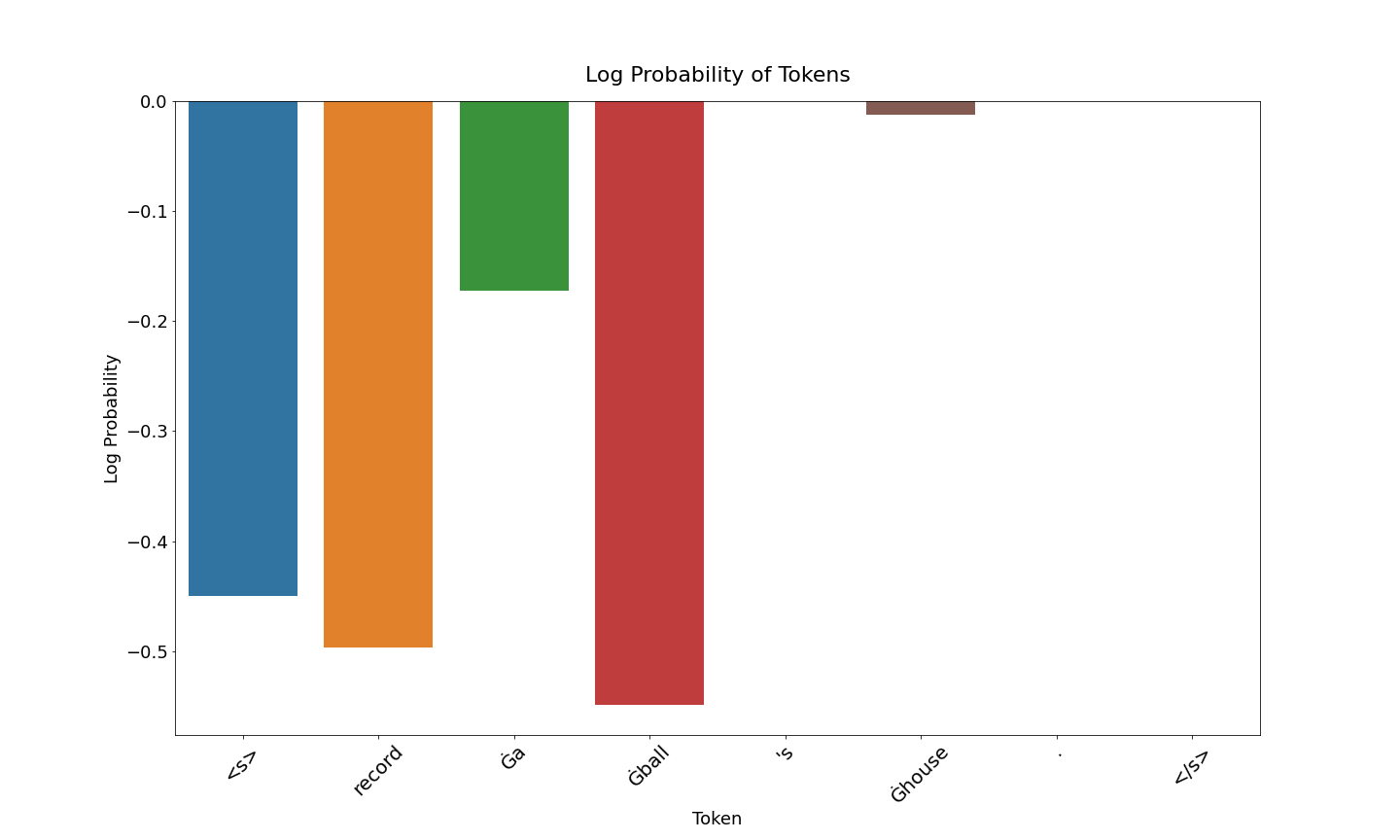}\label{}}
    \hfill
    \subfloat[Cross-attention]{\includegraphics[width=0.5\textwidth]{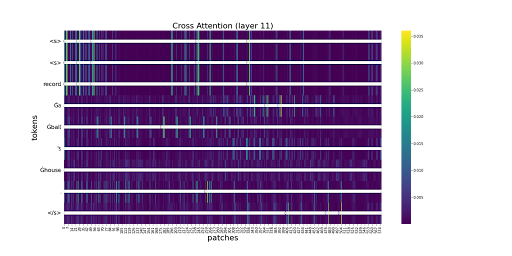}\label{fig:sub2}}
    \caption{Log probabilities and cross-attention for the transcripted tokens from Figure \ref{fig:lacuna}(f) with the following ground truth transcription: round a doll's house. }
    \label{fig:error}
\end{figure}

We investigate whether the attention entropy feature has a greater or lesser influence compared to our baseline log probability feature in identifying lacunas and characters with complex handwriting. We report our results related to attention in section \ref{section:attention}. The reason we train separate logistic regression models for detecting lacuna and other errors is that sometimes the lacuna characters get higher log probability and lower uniform attention when transcribed correctly. Conversely, we see higher attention and lower log probability in the scenario where non-lacuna tokens get mistranscribed due to other errors. This phenomenon makes it a hard problem to solve and thus we decide to train separate models to disentangle these cases. For instance, while transcribing the line image from Figure \ref{fig:lacuna}(f), we find that the model correctly transcribes the lacuna token \textit{house} and gives higher log probability (Figure \ref{fig:error}a) and lower uniform attention (Figure \ref{fig:error}b). Whereas for the non-lacuna tokens that are mistranscribed (such as \textit{record} and \textit{ball}), we see lower log probability and higher non-uniform attention for them.  

\subsection{Training TrOCR in Leiden Conventions}

The Leiden Conventions are adopted in many subfields of paleography to indicate the uncertainty of, and deviation from, the visual evidence for a transcription \cite{van_groningen_signis_1932}. Under these conventions, missing or uncertain information, such as illegible characters or words, should be denoted by enclosing them within square brackets. Characters transcribed with partial evidence should have dots placed under them.  For instance, using the Leiden conventions, the transcription for the line image in Figure \ref{fig:lacuna}(f) can be written as \textit{round a doll's h\textbf{[}o\textbf{]}use.}

We experimented with training TrOCR using the Leiden conventions in the ground truth transcriptions of line images. We annotated the transcriptions of 15\% of the line images containing a lacuna in our training data by enclosing the lacuna token within square brackets. We used a customized loss function to penalize the model for failing to generate square brackets in positions where it should enclose the lacuna within square brackets. This approach aimed to train the model to recognize and appropriately handle lacunae within the transcriptions using the Leiden conventions. Let $L$ be the original (cross-entropy) loss function. The custom weighted loss function $L_{\text{weighted}}$ is defined as:

\begin{center}
    \( L_{\text{weighted}} = L \times (n_{\text{brackets}} + 1) \) \\
\end{center}

\noindent where ($n_{\text{brackets}}$+1) is the weight and $n_{\text{brackets}}$ is the number of occurrences of square brackets $[ ]$ in the ground truth transcriptions. This equation ensures that the loss for each sample is multiplied by a weight that is proportional to the number of square brackets in its corresponding ground truth transcription. Samples with more square brackets will have higher weights, encouraging the model to learn the convention of generating square brackets around lacuna tokens.

\section{Experiments}
\label{sec:experiments}

We use the Aachen’s partition of the IAM Handwriting dataset\footnote{https://github.com/jpuigcerver/Laia/tree/master/egs/iam} where the training data has 6161 lines from 747 documents, validation data has 966 lines from 115 forms, and test data has 2915 lines from 336 forms. We generate lacuna versions corresponding to all line images in our dataset. Next, we train the following variations of models using the TrOCR base model (334M parameters). 
\begin{enumerate}
    \item Model trained solely on all binarized non-lacuna images (total 6161 images). 
    \item Model trained on all binarized images and their corresponding lacuna variations (total 6161*2 images). 
    \item Model trained on all binarized images in addition to 30\% randomly selected lacuna images (total 8009 images).
    \item Model trained on all binarized images supplemented with 15\% randomly selected lacuna images (total 7085 images).
\end{enumerate}

The first model is trained to assess whether the Transformer-based model handles lacunae without exposure to any lacuna images. TrOCR is pre-trained without any lacuna variations, so we were interested in evaluating its ability to leverage pre-training knowledge for inferring lacunae. Furthermore, we experiment with different proportions of lacuna images in the training data to observe their impact on model performance. We use the following hyperparameters for training: \textit{learning rate} of 2e-5, \textit{weight decay} of 0.0001, \textit{batch size} of 4, and \textit{AdamW} optimizer. We train our models for 20 epochs on four A100 40GB GPUs and pick the best model (with the lowest validation character error rate).

We examine the attention patterns and log probabilities specifically from the last layer (layer 12) and the first head. The TrOCR base model comprises 12 layers, with each layer containing 16 attention heads. However, in this study, we concentrate solely on the final layer and a single head. While we acknowledge the potential benefits of conducting a comprehensive analysis across all layers and heads, we opt for this focused approach for the current investigation. 

\section{Results}
\label{sec:results}

\subsection{Visual Examples: Lacuna Transcription Improvement}
\begin{figure}[h]
\centering
\begin{minipage}{0.18\textwidth}
\end{minipage}
\begin{minipage}{1\textwidth}
\centering
    \includegraphics[width=0.7\textwidth]{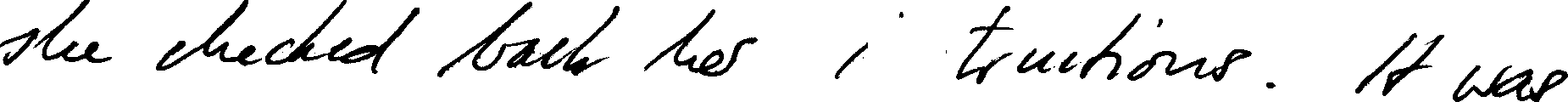}
\end{minipage}

\vspace{0.5cm}
\textbf{Ground truth}: she checked back her instructions. It was

\textbf{Non-lacuna model}: \textcolor{red}{the} checked back her \textcolor{red}{1 truth}. It was

\textbf{lacuna model}: she checked back her instructions. It was

\caption{Line image with a lacuna in \textit{instructions}. Errors are highlighted in red.}
\label{fig:r6}
\end{figure}

\begin{figure}[h]
\centering
\begin{minipage}{0.18\textwidth}
\end{minipage}
\begin{minipage}{1\textwidth}
\centering
    \includegraphics[width=0.7\textwidth]{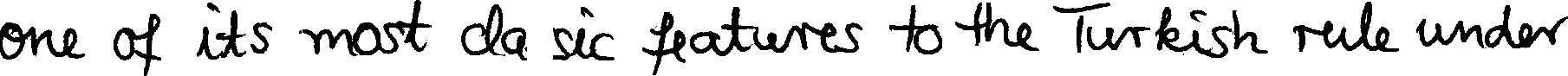}
\end{minipage}

\vspace{0.5cm}
\textbf{Ground truth}: one of its most classic features to the Turkish rule under

\textbf{Non-lacuna model}: one of its most \textcolor{red}{dc} \textcolor{red}{pictures} to the Turkish rule under

\textbf{lacuna model}: one of its most classic features to the Turkish rule under

\caption{Line image with a lacuna in \textit{classic}. Errors are highlighted in red.}
\label{fig:r5}
\end{figure}

\begin{figure}[h]
\centering
\begin{minipage}{0.18\textwidth}
\end{minipage}
\begin{minipage}{1\textwidth}
\centering
    \includegraphics[width=0.7\textwidth]{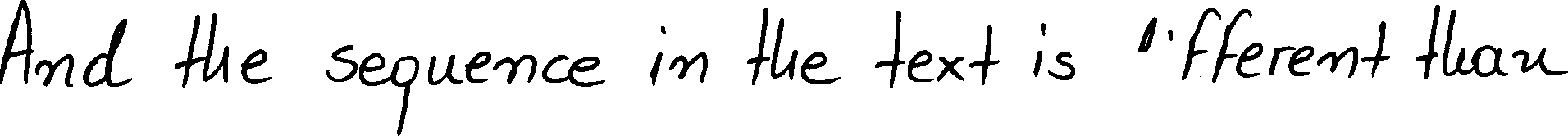}
\end{minipage}

\vspace{0.5cm}
\textbf{Ground truth}: And the sequence in the text is different than

\textbf{Non-lacuna model}: And the sequence in the text is \textcolor{red}{"} \textcolor{red}{f} \textcolor{red}{frequent} than

\textbf{lacuna model}: And the sequence in the text is different than

\caption{Line image with a lacuna in \textit{different}. Errors are highlighted in red.}
\label{fig:r4}
\end{figure}

In this section, we present images containing lacunae that were initially mistranscribed by the model trained only on non-lacuna images. However, our trained model, which is specifically trained on images with lacunae, successfully transcribes them accurately. We observe in Figure \ref{fig:r6} and Figure \ref{fig:r4} that multiple characters from a word form a lacuna, while in Figure \ref{fig:r5}, a single-character lacuna is evident. Remarkably, the model trained on lacuna images accurately transcribes all these variations.

%IMAGE 2 

\vspace{0.2cm}

%IMAGE 3 

\subsection{Performance Comparison of our Various Models}

\noindent \textbf{Model trained only on binarized non-lacuna images}:
The model trained only on clean binarized line images gives a validation CER of 7.46\%, CER of 11.49\% on the test set of non-lacuna images, and CER of 13.26\% on a mixture of binarized non-lacuna and lacuna line images. Table~\ref{table:table1} shows the model's performance in predicting both the lacuna and non-lacuna characters in the dataset. We observe that the model accurately transcribes non-lacuna characters 77.82\% of the time, while it achieves correct transcriptions for lacuna characters only 5.6\% of the time. One potential explanation could be that during pre-training, the TrOCR model may not have encountered any text instances containing missing characters similar to lacunae.

\begin{table}[t]
    \centering
    \caption{Confusion matrix for predicting (non-)lacuna characters}
     \subfloat[Only non-lacuna training]{\label{table:table1}
\begin{tabular}{c|cc}
        & \text{Lacuna} & \text{Non-Lacuna} \\
        \hline
        \multirow{2}{*}{\text{Correct}} & 5.6\% & 77.82\% \\
        & (\text{157/2800}) & (\text{17969/23089}) \\
        \hline
        \multirow{2}{*}{\text{Incorrect}} & 94.4\% & 22.18\% \\
        & (\text{2643/2800}) & (\text{5120/23089}) \\
    \end{tabular}
    }
    \subfloat[Non-lacuna + 100\% lacuna training]{\label{table:table2}
    \begin{tabular}{c|cc}
        & \text{Lacuna} & \text{Non-Lacuna} \\
        \hline
        \multirow{2}{*}{\text{Correct}} & 65.85\% & 75.38\% \\
        & (\text{1844/2800}) & (\text{17406/23089}) \\
        \hline
        \multirow{2}{*}{\text{Incorrect}} & 34.15\% & 24.62\% \\
        & (\text{956/2800}) & (\text{5683/23089}) \\
    \end{tabular}
    }
    
    \subfloat[Non-lacuna + 30\% lacuna training]{\label{table:table30}
    \begin{tabular}{c|cc}
        & \text{Lacuna} & \text{Non-Lacuna} \\
        \hline
        \multirow{2}{*}{\text{Correct}} & 62.25\% & 76.95\% \\
        & (\text{1743/2800}) & (\text{17767/23089}) \\
        \hline
        \multirow{2}{*}{\text{Incorrect}} & 37.75\% & 23.05\% \\
        & (\text{1057/2800}) & (\text{5322/23089}) \\
    \end{tabular}
    }
    \subfloat[Non-lacuna + 15\% lacuna training]{\label{table:table4}
    \begin{tabular}{c|cc}
        & \text{Lacuna} & \text{Non-Lacuna} \\
        \hline
        \multirow{2}{*}{\text{Correct}} & 60.14\% & 80.35\% \\
        & (\text{1684/2800}) & (\text{18554/23089}) \\
        \hline
        \multirow{2}{*}{\text{Incorrect}} & 39.86\% & 19.65\% \\
        & (\text{1116/2800}) & (\text{4535/23089}) \\
    \end{tabular}
    }
\end{table}

\noindent \textbf{Model trained on all binarized images and their corresponding lacuna variations}: The model trained on clean images and their lacuna variations yields a validation CER of 10\% and test CER of 13.75\% on a mixture of non-lacuna and lacuna images. Table \ref{table:table2} shows the model’s performance in predicting both the lacuna and non-lacuna characters in the dataset. We observe a notable enhancement in model performance when training on lacuna images, increasing accuracy from 5.6\% to 65.85\%. Specifically, the trained model accurately identifies 1844 out of 2800 lacuna characters, a substantial improvement compared to the baseline model's mere 157 correct predictions. Here, we can assert that training the model on labeled lacuna data teaches it to correctly recognize and restore any lacunae in the text. Another noteworthy observation is that training on lacuna images has a slight impact on the model's accuracy with clean, non-lacuna characters, evidenced by a decrease from 77.82\% to 75.38\%. This tradeoff implies that while there's an improvement in recognizing lacuna characters, there's a minor compromise in accuracy for non-lacuna characters.

\noindent \textbf{Model trained on all binarized images in addition to 30\% randomly selected lacuna images }: The model trained on 30\% lacuna images gives a validation CER of 8.71\% and a test CER of 13.04\% on a mixture of lacuna and non-lacuna images. Table \ref{table:table30} shows the model’s performance in predicting both the lacuna and non-lacuna characters in the dataset. We observe a decline in the model's ability to predict lacuna characters compared to when it's trained on 100\% of lacuna images. However, its accuracy in predicting non-lacuna characters increases, leading to a lower overall test CER. 

\noindent \textbf{Model trained on all binarized images in addition to 15\% randomly selected lacuna images}: The model trained on 15\% lacuna images gives a validation CER of 7.83\% and test CER of 11.11\% on a mixture of lacuna and non-lacuna images. Table \ref{table:table4} shows the model’s performance in predicting both the lacuna and non-lacuna characters in the dataset. In line with the model trained on 30\% lacuna images, there is a decrease in the accuracy of predicting lacuna characters, while the accuracy for non-lacuna characters shows improvement. This suggests that reducing the proportion of lacuna images in the training data significantly limits the model's ability to learn about lacunae compared to training with 100\% of lacuna images in the dataset. Another intriguing observation is the increase in the model's accuracy in correctly predicting clean non-lacuna characters from 77.82\% (accuracy of a model trained solely on binarized non-lacuna images) to 80.35\%. This suggests that the inclusion of additional lacuna images in the training set contributes to enhancing performance on non-lacuna characters in line images. It is possible that initially, the model might incorrectly predict a clean non-lacuna character. However, exposure to multiple copies of the same image (lacuna and non-lacuna versions) could have assisted the model in transcribing that character correctly. 

\subsection{Results from our Log Probability and Attention Experiments} 
\label{section:attention}
\begin{figure}[t]
    \centering
    \subfloat[Detecting lacunae with log probability and entropy features]{
        \includegraphics[width=0.35\textwidth]{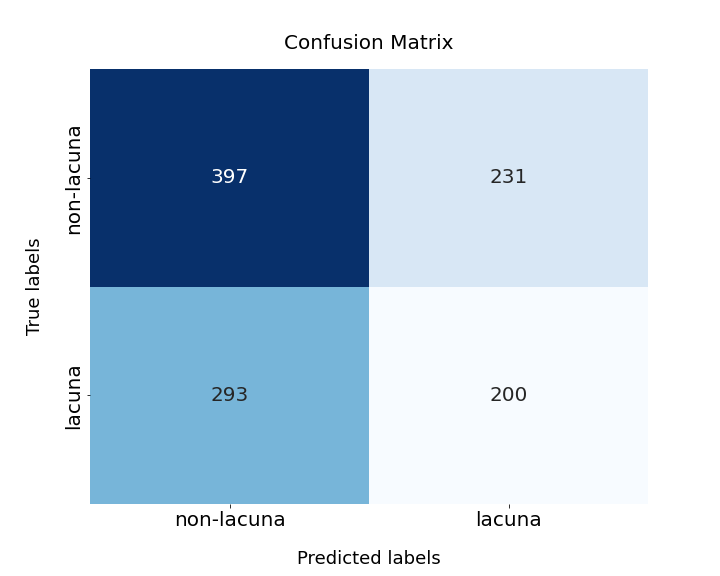}
        \label{fig:lacuna1}
    }\hfill 
    \subfloat[Detecting lacunae with log probability feature]{
        \includegraphics[width=0.35\textwidth]{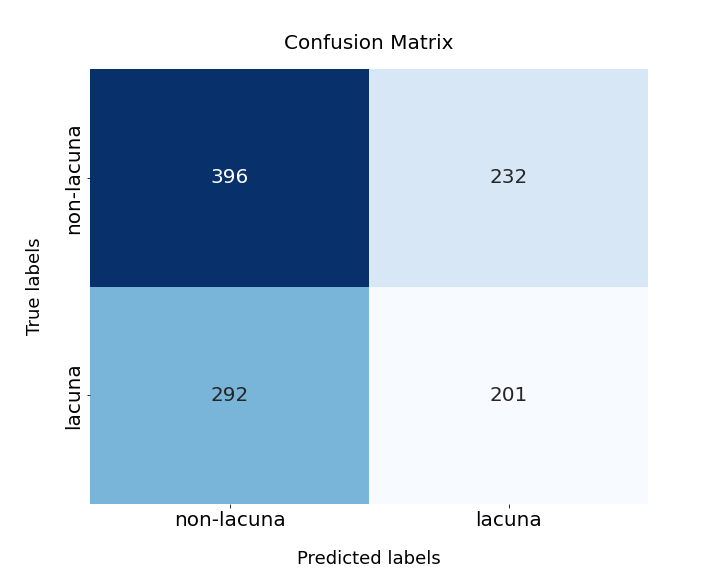}
        \label{fig:lacuna2}
    }
    
    \subfloat[Detecting errors with log probability and entropy features]{
        \includegraphics[width=0.35\textwidth]{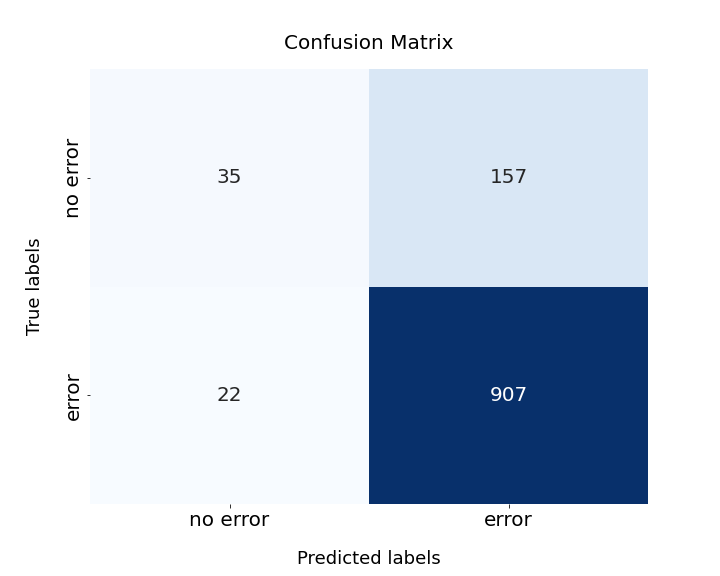}
        \label{fig:error1}
    }\hfill 
    \subfloat[Detecting errors with log probability feature]{
        \includegraphics[width=0.35\textwidth]{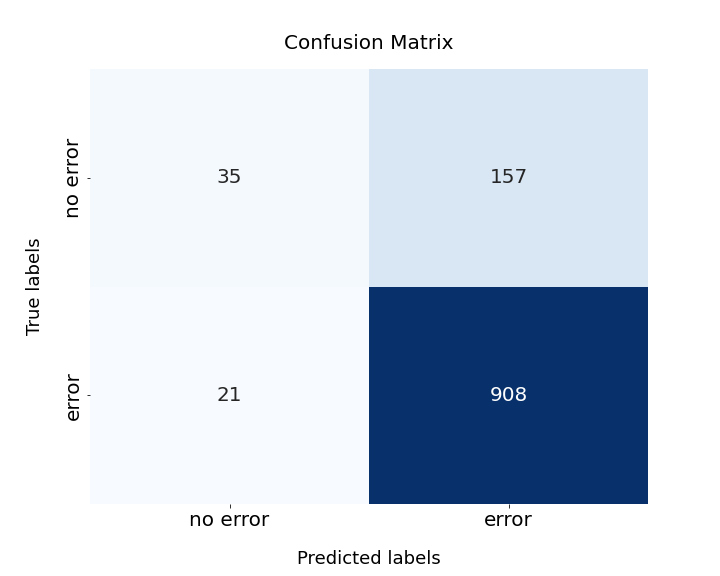}
        \label{fig:error2}
    }
    
    \caption{Confusion Matrices for logistic regression models trained to flag lacuna and errors in line images with log probability and attention entropy as features.}
    \label{fig:conf_matrices}
\end{figure}

Our findings indicate that attention does not outperform the log probability baseline in identifying lacunae and other errors in the line images. Table \ref{tab:detection} illustrates the efficacy of our logistic regression models trained on both log probability and attention entropy as features. We observe that attention entropy gets smaller feature importance coefficients compared to log probability in the classification of lacunae and transcription errors. The baseline models, using log probability as features, accurately predict whether a line image has any lacuna 53.25\% of the time and errors 84.12\% of the time. Figure~\ref{fig:conf_matrices} shows confusion matrices for all models trained to flag lacunae and errors in a line image. The model trained on log probability accurately identifies 396 non-lacuna images and 201 lacuna images. Similarly, it correctly flags 907 line images with transcription errors. Overall, log probability appears to be a robust metric to flag any lacunae and transcription errors in the line images. 

In our analysis, we focus exclusively on the last layer and the first head of attention. However, it is essential to explore whether these findings remain consistent across all layers and heads, as well as with different attention metrics. Such an investigation could provide valuable insights into the robustness and generalizability of our results. 

\begin{table}[tbp]
    \centering
     \caption{Comparison of log probability and attention entropy features shows no difference in predictive accuracy.}  
    \begin{tabular}{l p{6cm} c} 
        \toprule  
        Task & Feature Importance & Accuracy \\
        \midrule  
        Lacuna & Log probability: -0.247 & 53.25\% \\
        Lacuna & Log probability: -0.163 \& Entropy: 0.001 & 53.25\% \\
        Other errors & Log probability: -2.683 & 84.12\% \\
        Other errors & Log probability: -2.688 \& Entropy: 0.064 & 84.03\% \\
        \bottomrule  
    \end{tabular}
    \captionsetup{skip=10pt} 
    \label{tab:detection}  
\end{table}

\subsection{Results from training using Leiden conventions}
We observe that the model generates square brackets (\texttt{[]}) in the transcription but consistently fails to enclose lacuna characters within those brackets. For instance, instead of \texttt{round a doll's h[o]use} it generates \texttt{round a [] doll's house.} It may be worthwhile to investigate if increasing the proportion of training data containing transcriptions formatted with the Leiden conventions, coupled with using an alternative loss function, could enable the model to learn and generate the appropriate bracketing.  It may also be worth exploring the interaction of these conventions with the tokenization of the decoder model.  Putting brackets around whole output tokens, for instance, could still indicate the presence of a lacuna in a word without modifying output tokenization.

\section{Conclusion}
\label{sec:conclusion}

Transformer models like TrOCR can successfully learn to restore lacunae in line images when trained on datasets featuring such gaps. Log probability can be leveraged to flag lacunae and transcription errors (e.g., due to non-standard writing or inking issues). Our controlled experimental setup demonstrates the promising potential of transformer-based models for restoring damaged historical documents. Moreover, by studying mechanistic properties such as log probability, it is possible to infer whether a specific document or image contains damages or inconsistencies like lacunae or errors without explicitly looking at those documents or images. Looking at more fine-grained attention features over a wider range of model representations and model architectures would be an interesting area of future work.  Expanding our capacity for model interpretability could prove invaluable to scholars seeking to incorporate computational models into a paleographical workflow. Extending our methodology to real historical documents with lacunae will be the focus of our future work.
%to separate clean images from those with damages (such as lacunae) and errors.
 
%
% the environments 'definition', 'lemma', 'proposition', 'corollary',
% 'remark', and 'example' are defined in the LLNCS documentclass as well.
%

\newpage
 \bibliographystyle{splncs04}
 \bibliography{ref}

\begin{thebibliography}{10}
\providecommand{\url}[1]{\texttt{#1}}
\providecommand{\urlprefix}{URL }
\providecommand{\doi}[1]{https://doi.org/#1}

\bibitem{ArlitschHerbert+2004+59+67}
Arlitsch, K., Herbert, J.: Microfilm, paper, and ocr: Issues in newspaper digitization. the utah digital newspapers program  \textbf{33}(2),  59--67 (2004)

\bibitem{assael_restoring_2022}
Assael, Y., Sommerschield, T., Shillingford, B., Bordbar, M., Pavlopoulos, J., Chatzipanagiotou, M., Androutsopoulos, I., Prag, J., de~Freitas, N.: Restoring and attributing ancient texts using deep neural networks. Nature  \textbf{603}(7900),  280--283 (Mar 2022), publisher: Nature Publishing Group

\bibitem{NEURIPS2020_92d1e1eb}
Baevski, A., Zhou, Y., Mohamed, A., Auli, M.: wav2vec 2.0: A framework for self-supervised learning of speech representations. In: Larochelle, H., Ranzato, M., Hadsell, R., Balcan, M., Lin, H. (eds.) Advances in Neural Information Processing Systems. vol.~33, pp. 12449--12460. Curran Associates, Inc. (2020)

\bibitem{cowen-breen_logion_2023}
Cowen-Breen, C., Brooks, C., Haubold, J., Graziosi, B.: Logion: {Machine} {Learning} for {Greek} {Philology} (May 2023), arXiv:2305.01099 [cs]

\bibitem{devlin-etal-2019-bert}
Devlin, J., Chang, M.W., Lee, K., Toutanova, K.: {BERT}: Pre-training of deep bidirectional transformers for language understanding. In: Burstein, J., Doran, C., Solorio, T. (eds.) Proceedings of the 2019 Conference of the North {A}merican Chapter of the Association for Computational Linguistics: Human Language Technologies, Volume 1 (Long and Short Papers). pp. 4171--4186. Association for Computational Linguistics, Minneapolis, Minnesota (Jun 2019)

\bibitem{dong-smith-2018-multi}
Dong, R., Smith, D.: Multi-input attention for unsupervised {OCR} correction. In: Gurevych, I., Miyao, Y. (eds.) Proceedings of the 56th Annual Meeting of the Association for Computational Linguistics (Volume 1: Long Papers). pp. 2363--2372. Association for Computational Linguistics, Melbourne, Australia (Jul 2018)

\bibitem{van_groningen_signis_1932}
van Groningen, B.A.: De signis criticis in edendo adhibendis. Mnemosyne  \textbf{59}(4),  362--365 (1932), \url{https://www.jstor.org/stable/4426628}, publisher: Brill

\bibitem{ocr1}
H{\"a}m{\"a}l{\"a}inen, M., Hengchen, S.: From the paft to the fiiture: a fully automatic {NMT} and word embeddings method for {OCR} post-correction. In: Mitkov, R., Angelova, G. (eds.) Proceedings of the International Conference on Recent Advances in Natural Language Processing (RANLP 2019). pp. 431--436. INCOMA Ltd., Varna, Bulgaria (Sep 2019)

\bibitem{koenecke_careless_2024}
Koenecke, A., Choi, A.S.G., Mei, K., Schellmann, H., Sloane, M.: Careless {Whisper}: {Speech}-to-{Text} {Hallucination} {Harms} (Feb 2024), arXiv:2402.08021 [cs]

\bibitem{trocr}
Li, M., Lv, T., Chen, J., Cui, L., Lu, Y., Florencio, D., Zhang, C., Li, Z., Wei, F.: Trocr: Transformer-based optical character recognition with pre-trained models. Proceedings of the AAAI Conference on Artificial Intelligence  \textbf{37}(11),  13094--13102 (Jun 2023)

\bibitem{nockels_implications_2024}
Nockels, J., Gooding, P., Terras, M.: The implications of handwritten text recognition for accessing the past at scale. Journal of Documentation  \textbf{80}(7),  148--167 (Jan 2024). \doi{10.1108/JD-09-2023-0183}, \url{https://doi.org/10.1108/JD-09-2023-0183}

\bibitem{radford2019language}
Radford, A., Wu, J., Child, R., Luan, D., Amodei, D., Sutskever, I.: Language models are unsupervised multitask learners  (2019)

\bibitem{t5}
Raffel, C., Shazeer, N., Roberts, A., Lee, K., Narang, S., Matena, M., Zhou, Y., Li, W., Liu, P.J.: Exploring the limits of transfer learning with a unified text-to-text transformer. J. Mach. Learn. Res.  \textbf{21},  140:1--140:67 (2020)

\bibitem{rijhwani-etal-2020-ocr}
Rijhwani, S., Anastasopoulos, A., Neubig, G.: {OCR} {P}ost {C}orrection for {E}ndangered {L}anguage {T}exts. In: Webber, B., Cohn, T., He, Y., Liu, Y. (eds.) Proceedings of the 2020 Conference on Empirical Methods in Natural Language Processing (EMNLP). pp. 5931--5942. Association for Computational Linguistics, Online (Nov 2020)

\bibitem{rijhwani-etal-2021-lexically}
Rijhwani, S., Rosenblum, D., Anastasopoulos, A., Neubig, G.: Lexically aware semi-supervised learning for {OCR} post-correction. Transactions of the Association for Computational Linguistics  \textbf{9},  1285--1302 (2021)

\bibitem{vogler-etal-2022-lacuna}
Vogler, N., Allen, J., Miller, M., Berg-Kirkpatrick, T.: Lacuna reconstruction: Self-supervised pre-training for low-resource historical document transcription. In: Carpuat, M., de~Marneffe, M.C., Meza~Ruiz, I.V. (eds.) Findings of the Association for Computational Linguistics: NAACL 2022. pp. 206--216. Association for Computational Linguistics, Seattle, United States (Jul 2022)

\end{thebibliography}

%LNCS Homepage, \url{http://www.springer.com/lncs}. Last accessed 4
%\end{thebibliography}
\end{document}